\documentclass[journal]{IEEEtran}
\usepackage{cite}
\usepackage[pdftex]{graphicx}
\graphicspath{{../pdf/}{../jpeg/}{../figures/}}
\DeclareGraphicsExtensions{.pdf,.jpeg,.png}
\usepackage{amsmath}
\usepackage{algorithmic} 
\usepackage{array} 
\usepackage[caption=false,font=footnotesize]{subfig}
\usepackage{fixltx2e}
\usepackage{stfloats}
\usepackage{fancyhdr}

\begin{document}
\title{File Classification Based on \\ Spiking Neural Networks}

%
\author{\IEEEauthorblockN{Ana Stanojevic\IEEEauthorrefmark{1},
Giovanni Cherubini\IEEEauthorrefmark{2},
Timoleon Moraitis\IEEEauthorrefmark{3} and 
Abu Sebastian\IEEEauthorrefmark{4}}
\IEEEauthorblockA{\\IBM Research - Zurich,
R\"uschlikon, Switzerland\\
Email: \IEEEauthorrefmark{1}ans@zurich.ibm.com, \IEEEauthorrefmark{2}cbi@zurich.ibm.com, \IEEEauthorrefmark{3}timosmoraitis@gmail.com, \IEEEauthorrefmark{4}ase@zurich.ibm.com}}


\maketitle
\let\thefootnote\relax\footnotetext{© 2020 IEEE. Personal use of this material is permitted. Permission from IEEE must be obtained for all other uses, in any current or future media, including reprinting/republishing this material for advertising or promotional purposes, creating new collective works, for resale or redistribution to servers or lists, or reuse of any copyrighted component of this work in other works.}

\begin{abstract}
In this paper, we propose a system for file classification in large data sets based on spiking neural networks (SNNs). File information contained in key-value metadata pairs is mapped by a novel correlative temporal encoding scheme to spike patterns that are input to an SNN. The correlation between input spike patterns is determined by a file similarity measure. Unsupervised training of such networks using spike-timing-dependent plasticity (STDP) is addressed first. Then, supervised SNN training is considered by backpropagation of an error signal that is obtained by comparing the spike pattern at the output neurons with a target pattern representing the desired class. The classification accuracy is measured for various publicly available data sets with tens of thousands of elements, and compared with other learning algorithms, including logistic regression and support-vector machines. Simulation results indicate that the proposed SNN-based system using memristive synapses may represent a valid alternative to classical machine learning algorithms for inference tasks, especially in environments with asynchronous ingest of input data and limited resources.
\end{abstract}

\section{Introduction}
Today the exponential growth rate of data poses new and difficult challenges for data storage, management, and analytics. Labels attached to files describing their relevance or category provide significant assistance with the later handling and use of the data, for instance to optimize storage systems and policies \cite{IEEEISCAS:cognitive}, or to derive further insights by using the labelled datasets to train machine-learning algorithms. The growth rate of the data, and the associated level of resources it consumes, allow neither manual classification of the files, nor further resource-consuming classification systems. The main challenge can therefore be formulated as finding an efficient data classiﬁcation system for application in environments with limited resources.

\par Data is typically organized in files of various types (e.g. picture, text, etc.). Files are associated with metadata, available in terms of key-value pairs, consisting of metadata fields and their values. Here we consider file classification based on metadata, where by metadata we mean not only file-system information, but in general features that can be obtained by analyzing the content of a file \cite{IEEEISCAS:keyvalue_pairs}. Various models have been developed to find near-optimal solutions for data classification. The most powerful classification models today rely on neural networks. Artificial neural networks (ANNs) have achieved excellent results in many applications. However, ANNs that require high-precision arithmetic are in general inefficient in terms of power consumption. SNNs \cite{IEEEISCAS:SNNs}\cite{IEEEISCAS:SNNs_intro} rely on sequences of spikes (ones and zeros) rather than continuous values for neuronal communication and are thus significantly more efficient than other ANNs\cite{IEEEISCAS:SNN_efficiency}. Moreover, SNNs are particularly attractive when inputs are sparse and asynchronous, and when learning must be on-line and lifelong. This is the case in data management systems, as the files to be classified are often asynchronously ingested. Furthermore, efficient temporal encoding schemes could sparsify large streams of files, while in certain cases the learning must continue for the classification of new files.

\par In this paper, we propose a file classification system using SNNs, where the relevant information contained in key-value metadata pairs is mapped by a novel correlative temporal encoding scheme to spike patterns. The key-value pairs representation together with the proposed encoding scheme make this system well suited for handling the issue of different file types at the input. We investigate possible trade-offs between classification accuracy and implementation complexity. Firstly, we consider unsupervised or supervised training based on spike-timing-dependent plasticity (STDP), a biologically plausible rule that allows efficient learning. Then we focus on backpropagation-based supervised training, where the error signal is obtained by comparing the spike pattern at the output neurons with a target pattern representing the desired class. Furthermore, we introduce an efficient overall system architecture relying on memristive crossbar arrays for the realization of neuronal connectivity. Simulation results are presented for various publicly available data sets. The accuracy obtained with file classifier SNNs and synaptic weights realized by memristive crossbar arrays is compared with that achieved by other learning algorithms, including logistic regression and support-vector machines (SVM).

\section{System Architecture}
Figure 1 illustrates the proposed overall file classification system based on SNNs. We consider an input file set $\mathcal{D}$, where each element from $\mathcal{D}$ is described with \textit{K} feature values $[f_1, f_2, ... f_K]$, also referred to as key-value pairs, from a set $\mathcal{S}$. The objective is to classify each file in $\mathcal{D}$ to one of \textit{M} classes from a set $\mathcal{C} = \{c_1, c_2, ... c_M\}$. 
In general, each feature can be either categorical or numerical, whereas classes are discrete and do not imply an order. The classification task consists in inferring the class of an element in $\mathcal{D}$ given its feature values. 


\begin{figure*}[!t]
\centering
\subfloat{\includegraphics[width=18cm]{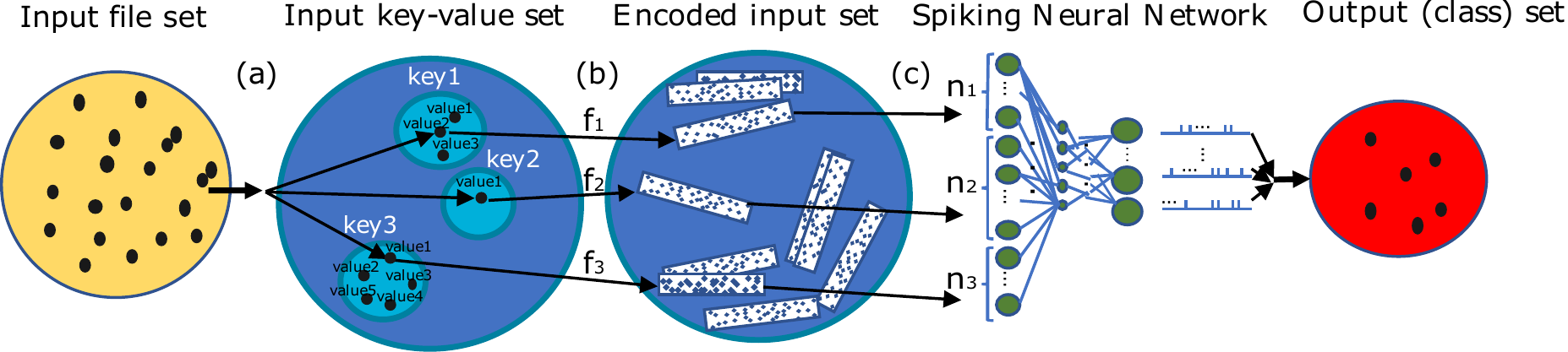}%
\label{fig_system}}
\caption{Architecture of the proposed file classification system based on SNNs. Information from the input files of the input set $\mathcal{D}$ is obtained in two steps: (a) mapping input files to the set of key-value pairs and (b) encoding of key-value pairs to spike patterns using a correlative time encoding (CTE) scheme. The encoded input spike pattern is then (c) sent to the input of an SNN which classifies the file to one of the \textit{M} classes in set $\mathcal{C}$. 
}
\end{figure*}

\subsection{Correlative Time Encoding (CTE)}
\begin{figure}[!thbp]
\centering
\includegraphics[width=8.7cm]{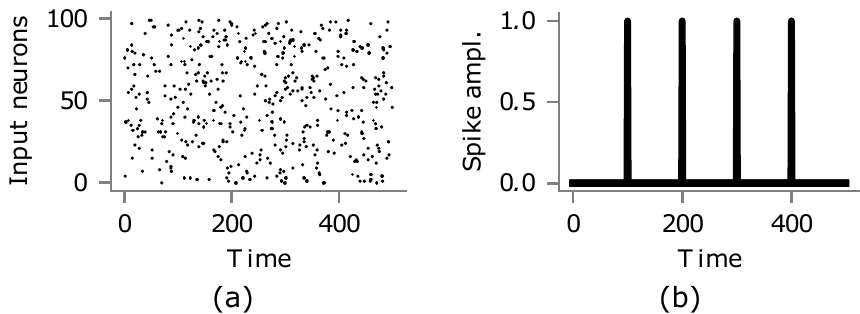}
\caption{An example of encoding in SNN: (a) input spike pattern for all input neurons and (b) output spike pattern for one output neuron.}
\label{fig_temporal_encoding}
\end{figure}

The spike communication brings energy efficiency to SNNs, however it is necessary to encode input files into input spike patterns and to decode output spike patterns into classes (Figure \ref{fig_temporal_encoding}). The information from each input file can be encoded into the rate of neuron spiking (\textit{rate encoding}) or into the precise timing of action potential (\textit{temporal encoding}) \cite{IEEEISCAS:thesisEncoding} (Figure 2(a)). The SNN produces an output spike pattern (Figure 2(b)). In case the output of the SNN is \textit{rate encoded} the class is predicted from a neuron that spikes with highest rate. In case of \textit{time encoded} output, for each class the target output pattern is predetermined. During training we attempt to induce spikes at those precise instants. The class is predicted as target spike pattern which is closest to the output spike pattern, using the \textit{van Rossum} distance metric \cite{IEEEISCAS:vanRossum}.
\par In \cite{IEEEISCAS:gardner} the input spike pattern was generated using \textit{temporal encoding}, where every neuron generates spikes as a random Poisson process. It was demonstrated that with such input encoding, the trained SNNs are able to reliably classify around 200 input spike patterns with 10 output classes. Other temporal encoding schemes were presented in \cite{IEEEISCAS:CTEbackground}\cite{IEEEISCAS:CTEbackground1}\cite{IEEEISCAS:CTEbackground2}. 
\par When generating temporally encoded input spike patterns, we would like to translate similarities among the input files into the  correlation of input spike patterns. Under the assumption that similar input files are often sharing the same class, this should facilitate the learning task for a neural network and improve the classification accuracy for input data sets that are larger than a few hundred files.  
Therefore, we propose a correlative temporal encoding (CTE) scheme, described as follows:
\begin{itemize}
\item for a given input file, metadata keys and corresponding values
are identified; 
\item each key $k$ is associated with a set of input neurons $n_k$ of the SNN; those neurons generate spikes at pseudo-random time instants, depending on the value the key assumes for the given input file; 
\item the input spiking neurons are fixed for a given key and pseudo-random spike patterns are fixed for a given key-value pair.
\end{itemize}
\begin{figure}[!thbp]
\centering
\includegraphics[width=8.7cm]{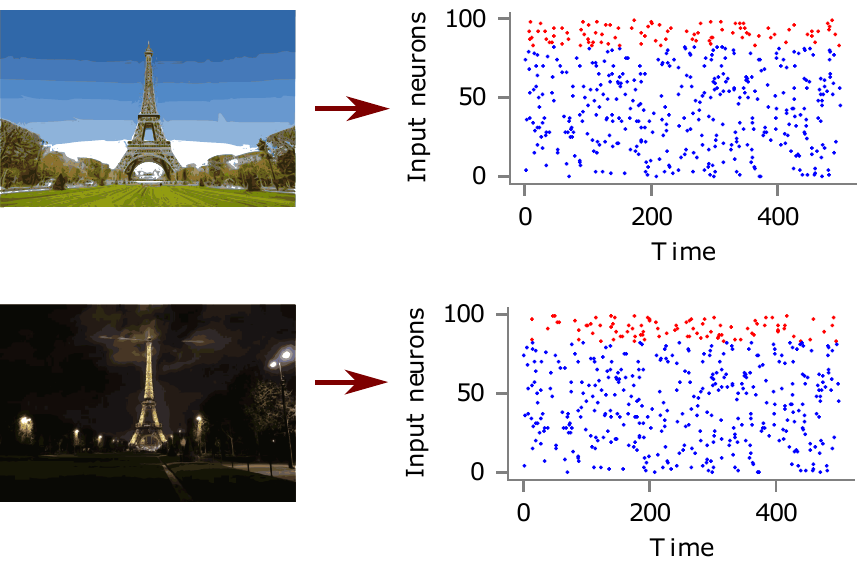}
\caption{An example of encoding two image input files using proposed CTE scheme. The two images are described with key-value pairs which are identical (both pictures are taken \textit{the same day}, of \textit{the same object}, etc.) except the one key-value pair describing the time of the day the picture was taken (\textit{day} or \textit{night}). Identical pseudo-random spike patterns correspond to the same key-value pairs (neurons 0-80). The neurons generating the input for the key describing the time of the day (neurons 81-100) are going to generate different pseudo-random spike patterns (red points).}
\label{fig_cte_example}
\end{figure} In CTE, same key-value pairs imply the same spike patterns. Therefore, the similarity among input files due to the same key-value pairs is translated into the correlation (inner product) between input spike patterns. An example of CTE for two input image files is given in Figure \ref{fig_cte_example}.

\subsection{Learning in SNNs}
We investigate three learning algorithms in conjunction with CTE to train file classifier SNNs: the STDP learning rule \cite{IEEEISCAS:STDP}\cite{IEEEISCAS:STDPhebb}\cite{IEEEISCAS:inhibition}\cite{IEEEISCAS:STDPdiehl} and two variants of the backpropagation algorithm \cite{IEEEISCAS:backpropagation}\cite{IEEEISCAS:SNN_deeplearning}\cite{IEEEISCAS:SNNs_backprop1}, one probabilistic and one deterministic. 
For STDP we assume the leaky-integrate-and-fire (LIF) neuron model and rate encoded output, while both backpropagation algorithms rely on time encoded target spike patterns. \\
Probabilistic backpropagation proposed in \cite{IEEEISCAS:gardner} assumes a spike response neuron model (SRM) \cite{IEEEISCAS:SRM} (a generalization of the LIF model). A neuron fires with probability $\rho(u)=1/(1+e^{(-\beta u)})$, given membrane potential \textit{u} and parameter $\beta$. The membrane potential at time instant $t$ of a postsynaptic neuron $j$ that receives input signals from presynaptic neurons $n$ is: 
\begin{equation}
    u_j(t) = \sum_{n}{w_{jn}(\mathcal{Y}_n \ast \epsilon)(t) + (\mathcal{Z}_j \ast \kappa)(t)},
\end{equation}
where $w_{jn}$ is the synaptic weight between neurons $n$ and $j$, $(\mathcal{Y}_n \ast \epsilon)(t)$ and $(\mathcal{Z}_j \ast \kappa)(t)$ denote convolution between a spike sequence and a postsynaptic potential (PSP) kernel $\epsilon$ and reset kernel $\kappa$, respectively, and where a spike sequence is given by a sequence of delta functions, $\mathcal{Y}_n(t) = \sum_{f}{\delta(t - t_n^f)}$. During the training process, the log-likelihood of target output patterns given output spike patterns from the SNN is maximized.\\
Deterministic backpropagation is proposed in  \cite{IEEEISCAS:snu}, where the neuron behavior is described by two discrete-time equations: 
\begin{equation}
\label{eq:SNU_LIF_DISCR}
\begin{array}{l}
    s_t = g(Wx_t + l(\tau)\odot s_{t-1}\odot (1-y_{t-1}))\\
    y_t = h(s_t + b),
  \end{array}
\end{equation}
where $x_t$ is the input, $s_t$ is  the  vector  of  internal  state  variables and $y_t$ is the output vector. Furthermore, $W$ is a synaptic weight matrix, $l(\tau)$ is a \textit{leaky} parameter, \textit{g} and \textit{h} are activation functions, $b$ is a bias implementing the spiking threshold and $\odot$ is the element-wise product. It is shown in \cite{IEEEISCAS:snu} that such spiking neural unit (SNU) is mimicking the behaviour of the LIF model. During the training at each time step $t$ the binary cross-entropy loss between the neuron output and the target signal representing the desired class is minimized. 
\begin{figure*}[!thbp]
\centering
\subfloat{\includegraphics[width=16cm]{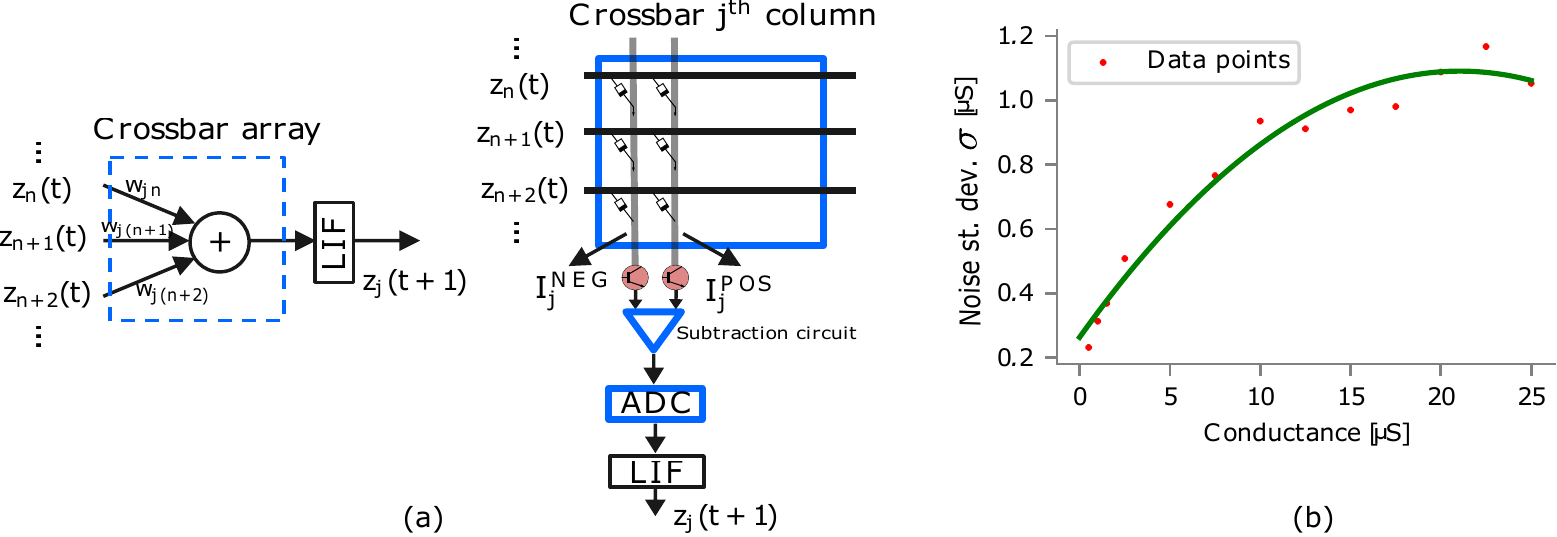}%
\label{fig_crossbar}}
\caption{(a) Neuronal connectivity implemented using memristive crossbar array, and (b) the conductance noise as a function of the programmed conductance value based on experimental measurements from 10,000 PCM devices from a prototype chip; the solid line represents the least squares fitting second-order polynomial to the data.}
\end{figure*}
\subsection {Implemented Models}
To investigate system performance and possible trade-offs with complexity of implementation, we consider the following CTE-based SNN models and learning techniques:
\begin{itemize}
    \item System 1: one-layer SNN with unsupervised or supervised STDP learning;
    \item System 2: one and two-layer SNNs, with probabilistic backpropagation \cite{IEEEISCAS:gardner};
    \item System 3: multi-layer SNNs, with determininstic backpropagation \cite{IEEEISCAS:snu}.
\end{itemize}

\subsection {Efficient Realization Using Memristive Synapses}
Recently, memristive devices \cite{Y2015wongNatureNano}\cite{Y2011chuaAPA} have been proposed as candidates for efficient realization of synaptic weights in ANNs and SNNs \cite{IEEEISCAS:memristive}\cite{IEEEISCAS:memristive1}. Memristive devices when organized in a crossbar array can be used to realize the neuronal connectivity in different layers of the SNNs, as shown in Figure 4(a). The devices represent the synaptic weights in terms of their conductance values and the synaptic efficacy can be emulated by exploiting the Ohm's and Kirchhoff's laws \cite{IEEEISCAS:memristive2}\cite{Y2013kuzumNanotechnology}. Typically, two devices organized in a differential configuration are used to represent the positive and negative synaptic weights \cite{Y2012bichlerTED}. Phase change memory (PCM) devices \cite{Y2016burrJETCAS} are arguably the most advanced memristive devices and are particularly well suited for synaptic realizations \cite{Y2018boybatNatComm}\cite{Y2016tumaEDL}, owing to their ease of programmability and low-power consumption. However, there are non negligible conductance variations associated with each programmed conductance value \cite{Y2019joshiArXiv}\cite{Y2019sebastianVLSI}. We experimentally characterized the achievable conductance values and the associated conductance variations for mushroom-type PCM devices fabricated in the 90nm technology node (see Figure 4(b)). In Section III, we will investigate the impact of these conductance variations on the performance of the file classification system.


\section{Simulation Results}
We select four data sets from the \textit{UCI Machine Learning Repository} for classification tasks \cite{IEEEISCAS:UCIcar}\cite{IEEEISCAS:UCInursery}\cite{IEEEISCAS:UCIincome}\cite{IEEEISCAS:UCIcf} to train and test the proposed system. 

\subsection{Baseline Models}
As baseline models, we consider standard logistic regression and SVM with a non-linear radial basis function (RBF) kernel. The two machine learning algorithms receive a non-temporal data to the input. The key-value pairs are categorical features (keys) represented as a one-hot encoded feature vector. 

\subsection{Comparative Study}
A comparison between the classification accuracy achieved by the three models introduced in Section 2 and the baseline algorithms is given in Table \ref{tab_software_results}.
\begin{figure}[!thbp]
\centering
\includegraphics[width=8.7cm]{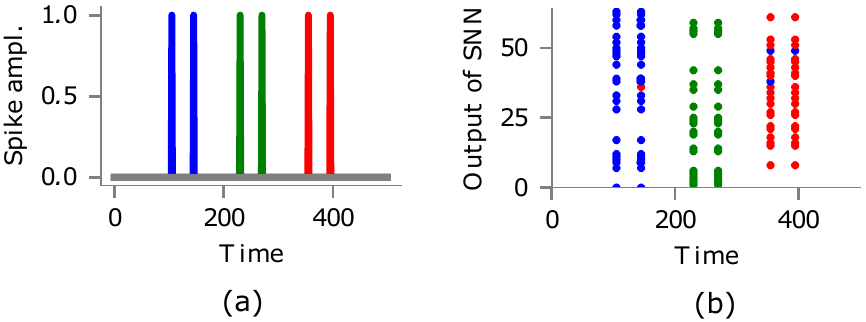}
\caption{(a) Target spike patterns for 3 classes and (b) output of the SNN for different input elements after training; classes are distinguished by color.}
\label{fig_snn_output}
\end{figure}

\begin{table}[!t]
\renewcommand{\arraystretch}{1.3}
\caption{Simulation Results for Different Data Sets.}
\label{tab_software_results}
\centering
\begin{tabular}{|c||c||c||c||c|}
\hline
Model&Adult&Nursery&Car&Connect\\
&Income&School&Evaluation&Four\\
\hline
Log. Reg.  &85.2\%& 93\%& 86.3\%& 75.6\%\\
\hline
SVM (RBF) &85.3\%&98.8\%&92.2\%&77.7\%\\
\hline
STDP unsup. &&&&\\
rate encoded &77.5\%& 67.2\%& 50.5\%& 64.9\%\\
\hline
STDP unsup. &77.8\%& 70.3\%& 59.6\%& 65.8\%\\
\hline
STDP sup. &75.6\%& 73.7\%& 70.2\%& 65.8\%\\
\hline
Probab. BackProp. &&&&\\
one layer &72.9\%& 73.9\%& 65.5\%& 59.8\%\\
\hline
Probab. BackProp. &&&&\\
two layers &79.2\%& 88.4\%& 87.7\%&64.5\%\\
\hline
Determ. BackProp. &&&&\\
two layers &85.4\%& 99.6\%& 96.7\%&74.4\%\\
\hline
Determ. BackProp.  &&&&\\
five layers &85.7\%& 99.3\%& 97\%&73.6\%\\
\hline
\end{tabular}
\end{table}
\fnbelowfloat

The high efficiency of the unsupervised or supervised STDP in \textit{System 1} may justify a trade-off in performance compared to the backpropagation-based algorithms. Moreover, the classification accuracy achieved even via unsupervised learning is respectable enough to envisage a continuously learning file classifier based on SNNs. For \textit{System 2}, the performance of a two-layer model is significantly better than the one-layer model, even though the derivation of the weight update rule involves larger approximations \cite{IEEEISCAS:gardner}. \textit{System 3} shows the best accuracy overall. This is thanks to the update rule obtained by the adopted backpropagation through time \cite{IEEEISCAS:snu}, which does not require the approximations of the probabilistic approach. Figure 5 shows the target spike patterns for \textit{System 3} and the achieved outputs. The results obtained for that system are close to or even surpassing logistic regression and non-linear SVM performance. 
\par We remark that the ability to obtain high accuracy for data sets with tens of thousands of elements was only possible owing to the introduction of the CTE scheme.

\subsection{Simulation Results on Memristive Synaptic Realization}
In order to evaluate the performance of the proposed system in hardware we resort to a hardware simulator. In particular, we focus on the two-layer model trained with deterministic backpropagation, which gives good performance with relatively low complexity. 
With reference to Figure 4, assuming the model has already been trained in software and stored, the implementation of the hardware simulator includes:
\begin{itemize}
    \item scaling the weights to conductance range [0$\mu$S, 25$\mu$S] 
    \item scaling the output of a crossbar to 8-bits ([-127, 127])
    \item scaling the 8-bit value back to the ranges that were obtained from the training data 
    \item adding the noise that originates from different sources to the conductance values (weights)
\end{itemize}
\par In Table \ref{tab_results_hardware} the results of inference on the hardware simulator are given, for various values of the conductance-dependent noise standard deviation. It can be seen that there is hardly any drop in classification accuracy when implemented using existing PCM devices. Moreover, these networks are found to be robust to even much higher levels of conductance variations. 

\begin{table}[!thbp]
\renewcommand{\arraystretch}{1.3}
\caption{Inference Results with Hardware Simulator.}
\label{tab_results_hardware}
\centering
\begin{tabular}{|c||c||c||c||c|}
\hline
Model&Adult&Nursery&Car&Connect\\
&Income&School&Eval.&Four\\
\hline
Software &85.4\%& 99.6\%& 96.7\%&74.4\%\\
\hline
HW (st. dev. = $\sigma$) &85.1\%& 99.4\%& 97\%&73.5\%\\
\hline
HW (st. dev. = 5 * $\sigma$)&81.6\%& 93.8\%& 93.2\%&69.6\%\\
\hline
HW (st. dev. = 10 * $\sigma$) &70.6\%& 73.1\%& 80.9\%&44.4\%\\
\hline
\end{tabular}
\end{table}
\section{Conclusion}
This paper introduces an SNN-based system to perform file classification with limited resources. We investigated different learning algorithms, models and configurations and evaluated their classification accuracy. We proposed a novel CTE scheme to map input elements that share same key-value pairs in the original data set to correlated input patterns, which leads to significantly improved performance. The classification accuracy achieved by deterministic backpropagation together with CTE is found to be comparable to, and even surpassing, that of logistic regression and non-linear SVM. In addition, we simulated a full system including hardware synapses implemented as memristive devices. Simulation results indicate that the hardware implementation of the system using memristive arrays wouldn't significantly affect performance. The presented work suggests that real time classification of asynchronously injested files could be a promising application for SNNs, especially if implemented efficiently using memristive hardware.


\newpage


%
\bibliographystyle{IEEEtran}
\bibliography{./bibtex/IEEEfull,./bibtex/IEEEabrv, ./bibtex/IEEEISCAS}
\end{document}